%% file: sideinfo.tex
\documentclass{article}
\usepackage{ijcai18}

\usepackage{times}
\usepackage{epsfig}
\usepackage{graphicx}
\usepackage{amsmath}
\usepackage{amssymb}
\usepackage{mathtools}
\usepackage{bbm}

\usepackage{natbib}

\usepackage{algorithm}
\usepackage{algorithmic}

\usepackage{hyperref}
\usepackage{url}

\title{Improving One-Shot Learning through Fusing Side Information}

\author{Yao-Hung Hubert Tsai$^{\dagger}$\,\,\,\,\,Ruslan Salakhutdinov$^{\dagger}$\\
$^{\dagger}$School of Computer Science, Machine Learning Department, Carnegie Mellon University\\
{\tt\small \{yaohungt, rsalakhu\}@cs.cmu.edu}
}

\begin{document}

\maketitle
\begin{abstract}
  Deep Neural Networks (DNNs) often struggle with one-shot learning where we have only one or a few labeled training examples per category. In this paper, we argue that by using side information, we may compensate the missing information across classes. We introduce two statistical approaches for fusing side information into data representation learning to improve one-shot learning. First, we propose to enforce the statistical dependency between data representations and multiple types of side information. Second, we introduce an attention mechanism to efficiently treat examples belonging to the `lots-of-examples' classes as quasi-samples (additional training samples) for `one-example' classes. We empirically show that our learning architecture improves over traditional softmax regression networks as well as state-of-the-art attentional regression networks on one-shot recognition tasks.
\end{abstract}

\input{intro}
\input{related}
\input{methods}
\input{regre}
\input{experiments}

\section{CONCLUSION}
{
	In this paper, we show how we can fuse multiple types of side information for better transferring knowledge across `lots-of-examples' classes and `one-example' classes to improve one-shot learning. Our contributions lie in two parts: (1) enforcing dependency maximization between learned image representations and learned label-affinity kernel, and (2) performing an attention mechanism for generating quasi-samples for `one-example' classes. 

	The form of side information can either be supervised/ unsupervised class embeddings or tree-based label hierarchy. We empirically evaluate our proposed method on both general and fine-grained datasets for one-shot recognition. The results consistently improve over traditional softmax regression model and the attentional regression model, which represents the current state-of-the-art for the one-shot learning problem. 

}

\newpage
{
\small
\bibliographystyle{apalike}
\bibliography{ijcai18}
}

\end{document}


\title{Supplementary for \\
Improving One-Shot Learning through Fusing Side Information}

\author{Yao-Hung Hubert Tsai$^{\dagger}$\,\,\,\,\,Ruslan Salakhutdinov$^{\dagger}$\\
$^{\dagger}$School of Computer Science, Machine Learning Department, Carnegie Mellon University\\
{\tt\small \{yaohungt, rsalakhu\}@cs.cmu.edu}
}

\date{}
\maketitle

\section{EXAMPLE FOR COVARIANCE MATRIX OF $\mathsf{AwA}$ DATASET}
{
	As an example, Fig. \ref{fig:cov} shows construction of the tree covariance matrix designed for a randomly picked subset in {\em Animals with Attributes} ($\mathsf{AwA}$) dataset \citep{lampert2014attribute}.

	\begin{figure}[h]
	\centering
	\includegraphics[width=0.48\textwidth]{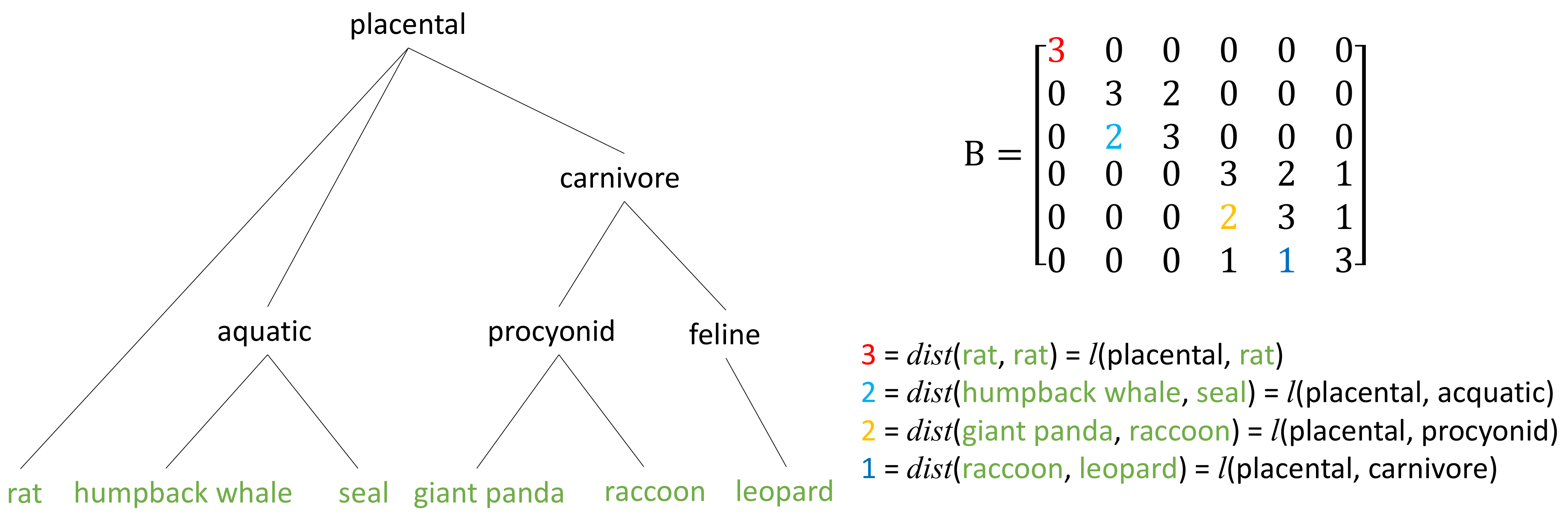}
	\caption{\footnotesize Tree structure and its corresponding tree covariance matrix inferred from {\em wordnet} structure for six randomly picked categories in $\mathsf{AwA}$ dataset. {\em dist($\cdot, \cdot$)} denotes the distance between two categories and {\em l($\cdot, \cdot$)} denotes the length of the branch between two nodes. Best viewed in color.}
	\label{fig:cov}
	\vspace{-3mm}
	\end{figure}
}

\section{REGRESSION}
{
	We adopt two regression strategies to form the label probability distributions. First, given the support set $\mathbf{S}$, we define the label prediction $\hat{y}$ to be
	\begin{equation}
		\hat{y} \coloneqq P_{\theta_X}({y}|{x}; \mathbf{S}).
	\label{eq:y_label_prob} 
	\end{equation}

	\subsubsection*{a) Softmax (Parametric) Regression:}

	Standard softmax regression has been widely used in deep networks such as VGG \citep{simonyan2014very}, GoogLeNet \citep{szegedy2015going}, and ResNet \citep{he2016deep}. The predicted label $\hat{y}$ can be written as 
	\begin{equation}
		\hat{y} = \mathrm{softmax}\,\big(\phi^\top g_{\theta_X}(x)\big),
	\label{eq:y_soft} 
	\end{equation}
	where $\phi$ represents the matrix that maps $g_{\theta_X}(x)$ to the label space in $\mathbf{S}$. Similarly, the predicted label $\hat{y}'$ of $x'$ under the support set $\mathbf{S}'$ would be $\hat{y}'\coloneqq P_{\theta_X}({y}'|{x}'; \mathbf{S}') = \mathrm{softmax}\,\big({\phi '}^\top g_{\theta_X}({x}')\big)$. Note that $\phi$ and $\phi'$ are different matrices.

	\subsubsection*{b) Attentional (Non-Parametric) Regression:}
	
	Attentional regression, proposed by \cite{vinyals2016matching}, represents state-of-the-art regression strategy for one-shot setting. The predicted label $\hat{y}$ over a data $x$ given the support set $\mathbf{S}$ is defined as
	\begin{equation}
		\hat{y}= \sum_{i \in \mathbf{S}} a_g(x, x_i) y_i,
	\label{eq:y_atten} 
	\end{equation}
	where $a_g(\cdot,\cdot)$ is the attention kernel on domains $\mathcal{G} \times \mathcal{G}$. In fact, this is a linear smoother \citep{buja1989linear} for non-parametric regression, with the choice of weight equal to $a_g(x, x_i)$. A possible design of this kernel is
	\begin{equation}
		a_g(x, x_i) = \frac{e^{k_g(x, x_i)}}{\sum_{j \in \mathbf{S}}e^{k_g(x, x_j)}}, 
	\label{eq:att}
	\end{equation} 
	which can also be viewed as an attentional memory mechanism in which $y_i$ acts as external memory and $a_g(\cdot,\cdot)$ computes merely the extent to which we retrieve this information according to the corresponding data $x_i$.

	Hence, $\hat{y} \coloneqq P_\theta({y}|{x}; \mathbf{S})$ can either be defined on softmax regression (eq. \eqref{eq:y_soft}) or attentional regression (eq. \eqref{eq:y_atten}). We note that using softmax regression requires learning an additional matrix (i.e., $\phi$), while the use of attentional regression requires the additional computation of traversing the data-label pairs in~$\mathbf{S}$.

}

	\begin{table*}[t]
	\footnotesize
	\centering
	\caption{\footnotesize Average performance (\%) for the different availability of side information.}
	\vspace{1mm}
	\label{tbl:avail}
	\scalebox{0.88}{
	\begin{tabular}{|c|c|c|c|c|c|c|c|}
	\hline
	\multicolumn{8}{|c|}{$\mathsf{CUB}$}                                \\ \hline \hline
	available side information & $\mathit{none}$ & $\mathit{att}$ & $\mathit{w2v}$ & $\mathit{glo}$ & $\mathit{hie}$ & $\mathit{att}$/$\mathit{w2v}$/$\mathit{glo}$ & $\mathit{all}$ \\ \hline
	$\mathsf{HSIC}$$_{softmax}$   & 26.93 $\pm$ 2.41  &  30.93 $\pm$ 2.25  & 30.67  $\pm$ 2.10 & 30.53 $\pm$ 2.42  & 32.15 $\pm$ 2.28  & 30.58 $\pm$ 2.12       &  31.49 $\pm$ 2.28 \\ 
	$\mathsf{HSIC}$$_{attention}$  & 29.12 $\pm$ 2.44   &  32.86 $\pm$ 2.34 &  33.37 $\pm$ 2.30 &  33.31 $\pm$ 2.50 &  {\bf 34.10 $\pm$ 2.40} & 33.72 $\pm$ 2.45  & 33.75  $\pm$ 2.43 \\ \hline \hline
	\multicolumn{8}{|c|}{$\mathsf{AwA}$}                                \\ \hline \hline
	available side information & $\mathit{none}$ & $\mathit{att}$ & $\mathit{w2v}$ & $\mathit{glo}$ & $\mathit{hie}$ & $\mathit{att}$/$\mathit{w2v}$/$\mathit{glo}$ & $\mathit{all}$ \\ \hline
	$\mathsf{HSIC}$$_{softmax}$  & 66.39 $\pm$ 5.38   &  70.08 $\pm$ 5.27 & 69.30 $\pm$  5.41 & 69.94 $\pm$  5.62 & 73.32 $\pm$ 5.12  &   70.44  $\pm$    6.74    &  71.29 $\pm$ 5.64 \\ 
	$\mathsf{HSIC}$$_{attention}$  & 72.27 $\pm$ 5.82   & 76.60 $\pm$ 5.05  & 76.60  $\pm$ 5.15 & {\bf 77.38  $\pm$ 5.15} &  76.88 $\pm$ 5.27 &   76.84   $\pm$  5.65     &   76.98 $\pm$ 4.99 \\ \hline
	\end{tabular}
	}
	\end{table*} 

	\begin{table}[t]
	\vspace{-2mm}
	\footnotesize
	\centering
	\caption{\footnotesize Average performance (\%) for generalized one-shot recognition tasks. Our proposed methods jointly learn with all four side information: $\mathit{att}$, $\mathit{w2v}$, $\mathit{glo}$, and $\mathit{hie}$.}
	\vspace{1mm}
	\label{tbl:expand}
	\scalebox{0.85}{
	\begin{tabular}{|c||c|c|}
	\hline
	network / Dataset             & $\mathsf{CUB}$           & $\mathsf{AwA}$   \\ \hline \hline
	$\mathsf{softmax\_net}$    &     6.33  $\pm$  1.20      &    2.58  $\pm$  1.81    \\ 
	$\mathsf{HSIC}$$_{softmax}^\dagger$            &    9.29     $\pm$  2.15     &   3.04 $\pm$  1.87         \\ \hline \hline
	$\mathsf{attention\_net}$ [\cite{vinyals2016matching}]                 &     9.37  $\pm$  1.72     &    18.92    $\pm$   6.42   \\ 
	$\mathsf{HSIC}$$_{attention}^\dagger$       &   {\bf 10.21    $\pm$  1.92}       &   {\bf 28.89    $\pm$  6.07 }     \\ \hline
	\end{tabular}
	}
	\end{table}

\section{IMPLEMENTATION DETAILS}
{

	First, we treat the learning of embeddings $g_{\theta_X}(x) = g_{\theta^{\circ}_X}(g_{GoogLeNet}(x))$, where $g_{GoogLeNet}(x)$ denotes the mapping before the last layer of GoogLeNet \citep{szegedy2015going} pre-trained on ImageNet \citep{deng2009imagenet} images. We fix $g_{GoogLeNet}(x)$ without fine-tuning, and therefore the learning of $g_{\theta_X}(\cdot)$ can be relaxed as the learning of $g_{\theta^{\circ}_X}(\cdot)$.

	For model parameters $\theta_X$, we parameterize $g_{\theta_X^{\circ}}(\cdot)$ as two-hidden layer fully-connected neural network with dimensions $1024-500-100$, where $1024$ is the input dimension of the input GoogLeNet features. $tanh$ is chosen to be our activation function and we adopt $l_2-$normalization after its output. For the softmax regression part, $\phi$/$\phi '$ are parameterized  as one-hidden layer fully-connected neural network with dimensions $100-C/C'$. Then, we parametrize the mapping $f_{t,\theta_R}(\cdot)$ for class embeddings to be a two-hidden layer fully-connected neural network with dimensions $d_{t}-d_{c}-50$, where $d_{t}$ is the input dimension of the class embeddings from $R^t$. We choose $d_c = 100$ when $d_t > 100$ and $d_c = 75$ when $d_t < 100$. We also adopt $tanh$ as the activation function and use $l_2-$normalization after its output.

	The trade-off parameter $\alpha$ is set to $0.1$ for all the experiments. To decide the value of $\alpha$, we first divide the 'lots-of-examples' classes into two splits (i.e., one for training and another for validation) and perform cross-validation on $\alpha$ from $10^-3$, $10^-2$, ..., $10^3$.

	In each trial, we fix $\mathbf{S}'$ to contain all `few-examples' classes and fix $|\mathbf{S}_{batch}| = 256$. The model is implemented in TensorFlow \citep{abadi2015tensorflow} with Adam \citep{kingma2014adam} for optimization. We observe that for {\em softmax regression}, the model converges within $500$ iterations; on the other hand, for {\em attentional regression}, the model converges within $100$ iterations. 
}

\section{AVAILABILITY OF VARIOUS TYPES OF SIDE INFORMATION}
{
	In Table \ref{tbl:avail}, we evaluate our proposed methods when not all four types of side information are available during training. It is surprising to find that there is no particular rule of combining multiple side information or using a single side information to obtain the best performance. A possible reason would be the non-optima for using kernel average in eq. (3). That is to say, in our current setting, we equally treat contribution of every type of side information to the learning of our label-affinity kernel. Nevertheless, we still enjoy performance improvement of using side information compared to not using it.

}

\section{REMARKS ON DIRECT SIDE INFORMATION FUSION}
{
	In the paper, we propose a method that fuses the side information indirectly, in which we enforce the dependency between the embeddings of class and data. Here, we examine the effect of the direct side information fusion. We conduct additional experiments by concatenating {\it att} attributes to image feature representations and then training the CNN classifier which is the exact one in our architecture. Followed by the same evaluation protocol, the average performance is 63.15\%. Our proposed method, on the other hand, achieves the accuracy of 70.08\%. One can also take into account both indirect and direct fusion for side information, which is part of our future work.
}

\section{COMPARISON WITH ReViSE \citep{tsai2017learning}}
{
	Here, we provide additional comparisons with ReViSE \citep{tsai2017learning}. Specifically, for each test class, we randomly label 3 images and train ReViSE together with the side information att. The average performance over 40 trials is 86.2\%. Our proposed method achieves 85.2\% which is comparable to ReViSE. 
}

\section{EXPANDING TRAINING- AND TEST-TIME CATEGORIES SEARCH SPACE}
{
	Another interesting experiment is to expand the training- and test-time search space to cover all training and test classes. While most of the one-shot learning papers \citep{santoro2016one,vinyals2016matching,ravi2017optimization} do not consider this setting, we consider it to be more practical for real-world applications. We alter the regression for both softmax and attentional version so that all classes are covered in the search space. In other words, the output label is now a vector of size $\mathbb{R}^{C+C'}$. 
	
	After expanding the categories' search space, it is meaningless to construct quasi-samples for `one-example' classes from samples in `lots-of-examples' classes. Therefore, we compare only $\mathsf{HSIC}$$_{softmax}^\dagger$ and $\mathsf{HSIC}$$_{attention}^\dagger$ with $\mathsf{softmax\_net}$ and $\mathsf{attention\_net}$.

	Table \ref{tbl:expand} shows the results of our experiment. First, a dramatic performance drop appears in every method compared to those that do not expand the search space. 
Objects in $\mathsf{CUB}$ and $\mathsf{AwA}$ all suffer from the confusion between training and test classes. 
Note that when considering one-shot setting, we have only one labeled data per test category during training time. Therefore, expanding the label search space makes the regression only focus on the `lots-of-examples' classes. 

}

\section{TREE HIERARCHY FOR DATASETS}
{
	Fig. \ref{fig:awa} is the tree hierarchy for Animal with Attributes ($\mathsf{AwA}$) dataset and Fig. \ref{fig:cub} is the tree hierarchy for Caltech-UCSD Birds 200-2011 ($\mathsf{CUB}$) dataset. The leaf nodes in the tree denote the class, and the internal nodes represent the super-class in {\em wordnet} structure.

	\begin{figure*}[h!]
	\centering
	\includegraphics[height=1.3\textwidth]{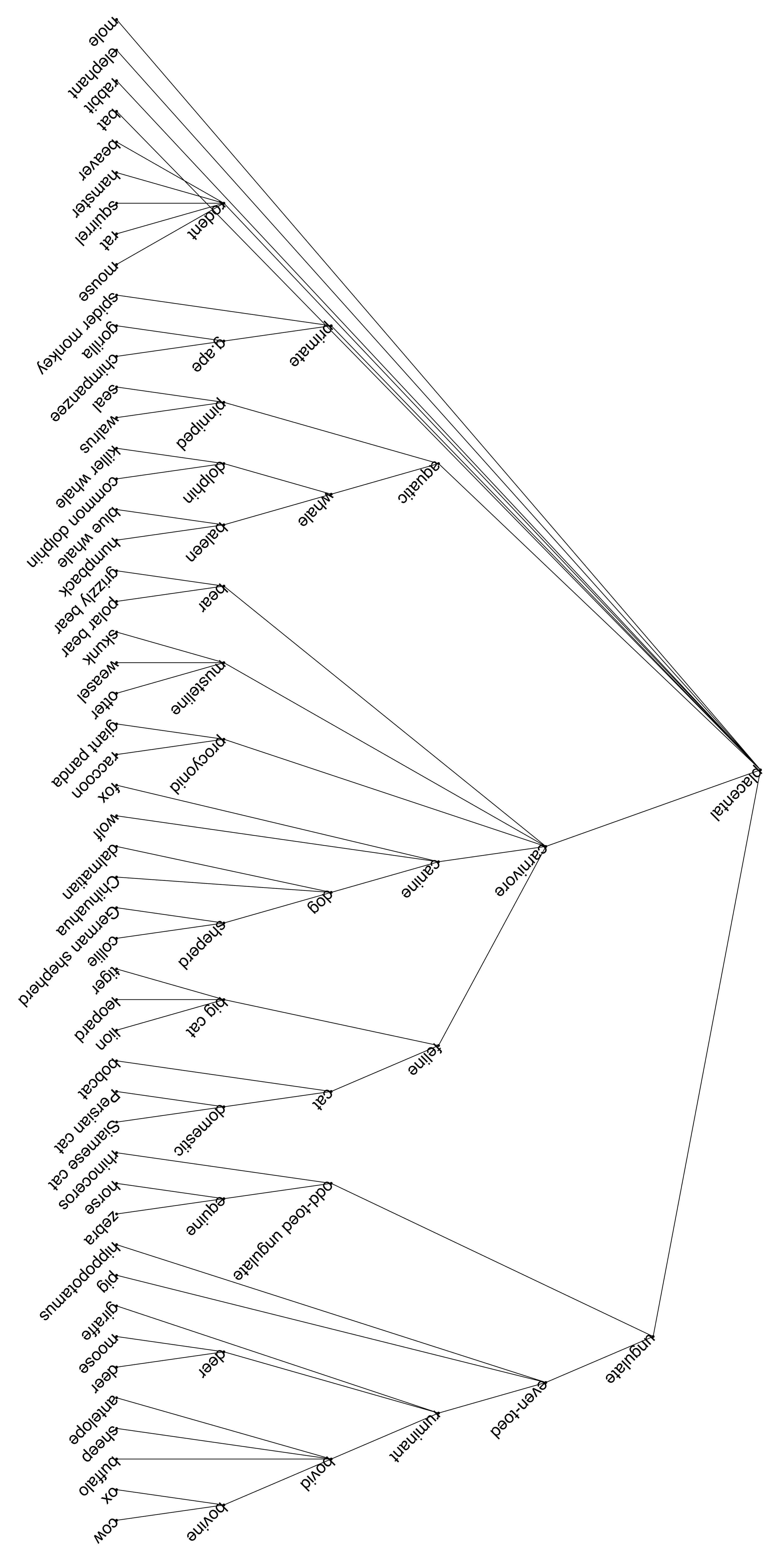}
	\caption{\footnotesize Tree hierarchy for Animal with Attributes ($\mathsf{AwA}$) dataset.}
	\label{fig:awa}
	\vspace{-3mm}
	\end{figure*}

	\begin{figure*}[t!]
	\centering
	\vspace{-5mm}
	\includegraphics[height=1.4\textwidth]{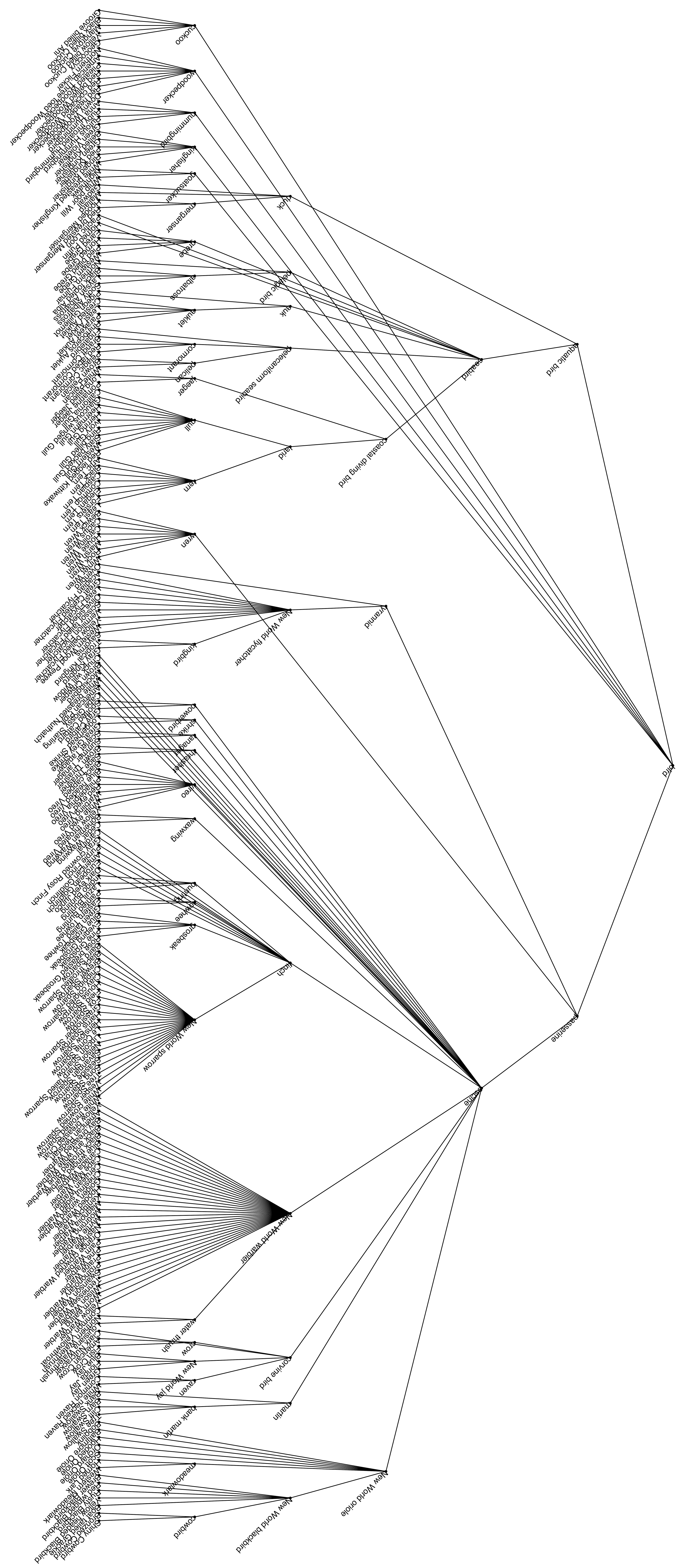}
	\caption{\footnotesize Tree hierarchy for Caltech-UCSD Birds 200-2011 ($\mathsf{CUB}$) dataset.}
	\label{fig:cub}
	\vspace{-3mm}
	\end{figure*}
}

{
\small
\bibliographystyle{apalike}
\bibliography{./ijcai18}
}

%% file: intro.tex
\section{Introduction}
\label{sec:intro}
{
	\begin{figure*}[t!]
	\centering
	\includegraphics[width=0.85\textwidth]{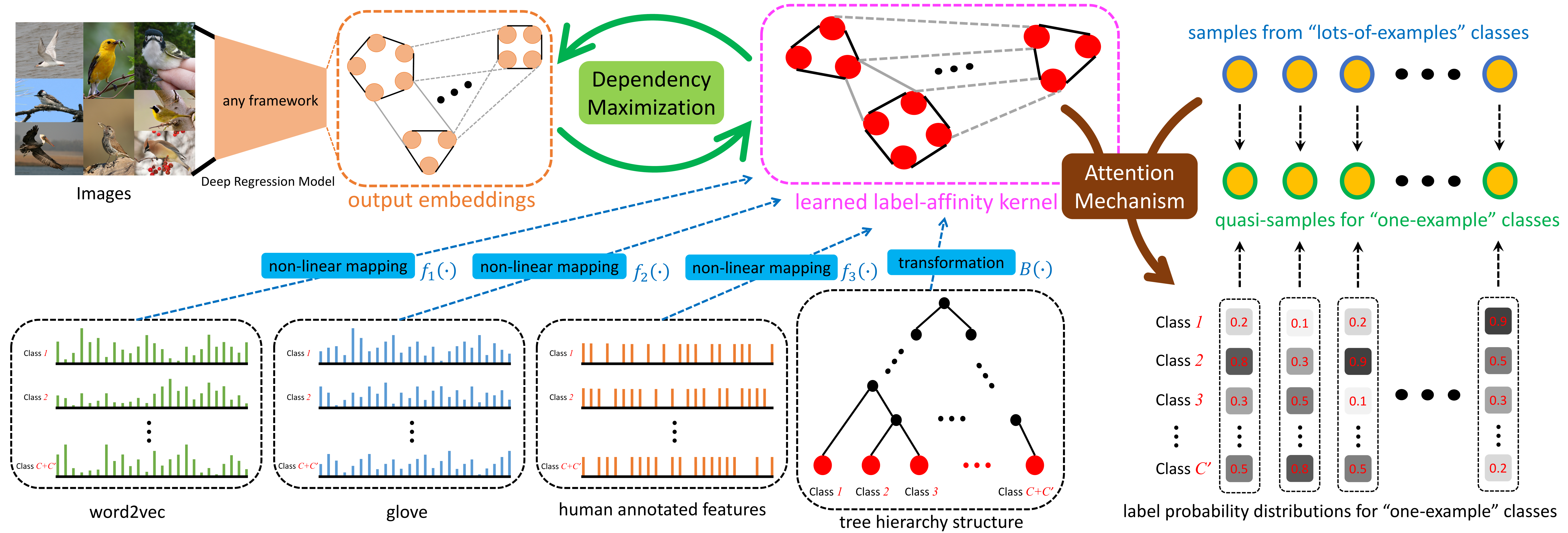}
	\caption{\footnotesize Fusing side information when learning data representation. We first construct a label-affinity kernel through deep kernel learning using multiple types of side information. Then, we enforce the dependency maximization criteria between the learned label-affinity kernel and the output embeddings of a regression model (it can either be the parametric softmax regression model or non-parametric attentional regression model \citep{vinyals2016matching}). Samples in `lots-of-examples' classes are used to generate quasi-samples for `one-example' classes. These generated quasi-samples can be viewed as additional training data.}
	\label{fig:illus}
	\end{figure*}

Training deep neural networks (DNNs) often requires lots of labeled examples, and they can fail to generalize well on new concepts that contain few labeled instances. Humans, on the other hand, can learn similar categories 
with a handful or even a single training sample~\citep{lake2015human}. In this paper, we focus on the extreme case: {\em one-shot learning} which has only one training sample per category. This `one-shot learning' ability has emerged as one of the most promising yet
challenging areas of research~\citep{lake2016building}. 


	We treat the problem of one-shot learning to be a transfer learning problem: how to efficiently transfer the knowledge from `lots-of-examples' to `one-example' classes. In the context of deep networks, 
one of the simplest transfer learning techniques 
is {\em fine-tuning} \citep{bengio2012deep}. However, fine-tuning may fail to work if the target task (e.g., regression on `one-example' classes) diverges heavily from the training task (e.g., regression on `lots-of-examples' classes)~\citep{yosinski2014transferable}. Alternatively, we can fuse side information for compensating the missing information across classes.

	In the paper, side information represents the relationship or prior knowledge between categories. For example, unsupervised feature vectors of categories derived from Wikipedia such as Word2Vec vectors \citep{mikolov2013distributed}, or tree hierarchy label structure such as WordNet structure \citep{miller1995wordnet}. In this work, we introduce two statistical approaches for fusing side information into deep representation learning.

	First, we propose to learn a label-affinity kernel from various types of side information. Our goal is to maximize Hilbert-Schmidt Independence Criterion (HSIC)~\citep{gretton2005measuring} between this kernel and the data representation embeddings. Since HSIC serves as a statistical dependency measurement, the learned data representations can be maximally dependent on the corresponding labels. Note that the label space spans over `lots-of-examples' to `one-example' classes, allowing us to bridge the gap between these categories.

	Second, to achieve better adaptation from `lots-of-examples' to `one-example' classes, we introduce an attention mechanism for `lots-of-examples' classes on the learned label-affinity kernel. Specifically,  we enable every sample in `lots-of-examples' classes to form a label probability distribution on the labels for `one-example' classes. Hence, each instance in `lots-of-examples' classes can be viewed as a quasi-sample for `one-example' classes and can be used as additional training data. 


In our experiments, we incorporate the proposed architecture in parametric softmax regression model and non-parametric attentional regression model introduced by \cite{vinyals2016matching}. We demonstrate improved recognition results on Animals with Attributes \citep{lampert2014attribute} and Caltech-UCSD Birds 200-2011 \citep{WelinderEtal2010} dataset. 
}

%% file: related.tex
\section{Related Work}
\label{sec:related}
{
There is a large body of research on transfer and one-shot learning.
Here, we focus on recent advances in {fusing side information} and {one-shot learning within deep learning}.

	{\bf Fusing Side Information:} \cite{srivastava2013discriminative} proposed to embed tree-based priors in training deep networks for improving objects classification performance. They enforced similar classes discovered from the tree-based priors to share similar weights of the last layer in deep networks. \cite{hoffman2016learning} presented a modality hallucination architecture for RGB image detection objective by incorporating depth of the images as side information. \cite{hoang2016incorporating} proposed to condition the recurrent neural network language models on metadata, such as document titles, authorship, and time stamp. For cross-lingual modeling, they also observed the improvement by integrating side information from the foreign language.

Many of the methods mentioned above attempt to indirectly strengthen the dependency between the side information and the learned data representations. Our approach, on the other hand, chooses to maximize this dependency directly under a statistical criterion.

	{\bf One-Shot Learning:} 
Deep learning based approaches to one-shot learning 
can be divided into two broad categories: {\em meta-learning approaches} and {\em metric-learning approaches}. On one hand, {\em meta-learning approaches} 
tackle the problem using a two-level-learning regime.
The first stage aims to quickly acquire knowledge of individual base tasks, while the second stage aims to extract meta-information from them. Memory-Augmented Neural Networks (MANN) \citep{santoro2016one} extended Neural Turing Machines for the meta-learning purpose so that they could rapidly bind never-seen information after a single presentation via external memory module. \cite{wood2016active} further extended MANN to learning to learn an active learner by using reinforcement learning. Different from other approaches, \cite{kaiser2017rare} approached one-shot learning problem in a life-long manner by introducing a long-term memory module. 
\citep{ravi2017optimization} proposed to learn the optimization algorithm for the {\em learner} neural network in the few-shot regime by an LSTM-based {\em meta-learner} model. More recent work~\citep{finn2017model,munkhdalai2017meta} embraced similar approaches with the goal of rapid generalization on few and never-before-seen classes.

On the other hand, {\em metric-learning approaches} choose to design a specific metric loss or develop a particular training strategy for one-shot learning. Deep Siamese Neural Networks \citep{koch2015siamese} designed a unique similarity matching criterion in deep convolutional siamese networks for one-shot image classification. Matching Networks (MN) \citep{vinyals2016matching} proposed a training strategy that aimed at training the network to do one-shot learning and also introduced an attentional regression loss to replace the standard softmax regression loss. Neural Statistician \citep{edwards2017toward} held a different viewpoint that a {\em machine learner} should deal with the datasets, instead of the individual data points. 
They developed an extension to the variational auto-encoders that can compute the {statistics} of a given dataset in an unsupervised fashion. Other recent work, including Skip Residual Pairwise Net (SRPN) \citep{mehrotra2017generative} and Prototypical Networks \citep{snell2017prototypical} lay in the same domain of {metric-learning approaches}.

Our approach can be easily incorporated into the {\em metric-learning} ones, as we detail in Sec.~\ref{sec:exp}. Instead of learning the networks exclusively from data, we extend the training from data and side information jointly. Since side information stands for the relationships between categories, we may compensate the missing information from `lots-of-examples' to `one-example' classes. 

%% file: methods.tex
\section{Proposed Method}
\label{sec:prop}
{
\subsection{Notation}
\label{subsec:joint_side}
{
	Suppose we have a support set $\mathbf{S}$ for the classes with lots of training examples. $\mathbf{S}$ consists of $N$ data-label pairs $\mathbf{S} = \{\mathbf{X}, \mathbf{Y}\}= \{x_i, y_i\}_{i=1}^{N}$ in which class $y_i$ is represented as a one-hot vector with $C$ classes. Moreover, we have $M$ different types of side information $\mathbf{R} = \{R^1, R^2, \cdots , R^M\}$, where $R^m$ 
can either be supervised/ unsupervised class embedding vectors or a tree-based label hierarchy, such as Wordnet \citep{miller1995wordnet}. Similarly, a different support set $\mathbf{S'}$ stands for `one-example' classes where $\mathbf{S'} = \{\mathbf{X'}, \mathbf{Y'}\} = \{x'_i, y'_i\}_{i=1}^{N'}$ in which class $y'_i$ is represented as a one-hot vector with $C'$ classes (disjoint from the classes in $\mathbf{S}$). $\mathbf{R'} = \{R'^1, R'^2, \cdots , R'^M\}$ then stands for the corresponding side information for $\mathbf{S'}$. Last, $\theta_X$ and $\theta_R$ are the model parameters dealing with the data and side information, respectively. 

One of our goals is to learn the embeddings of the data $g_{\theta_X}(x)$ ( $g_{\theta_X}(\cdot)$ denotes the non-linear mapping for data $x$ from $\{\mathbf{X},\mathbf{X'} \}$) that maximally align with the provided side information $\{\mathbf{R},\mathbf{R'}\}$. This can be done by introducing Hilbert-Schmidt Independence Criterion (HSIC) \citep{gretton2005measuring} into our architecture, as we detail in Sec.~\ref{subsec:joint_side}. 

In Sec. \ref{subsec:joint_side} and \ref{subsec:DKL}, for clarity of presentation,
we focus on learning dependency measure between $\mathbf{X}$ and $\mathbf{R}$. However, it can be easily extended to $\mathbf{X}'$ and $\mathbf{R}'$ or $\{\mathbf{X},\mathbf{X'} \}$ and $\{\mathbf{R},\mathbf{R'}\}$.

}
\subsection{Dependency Measure on Data and Side Information}
\label{subsec:joint_side}
{
	The output embeddings $g_{\theta_X}(\mathbf{X})$ and side information $\mathbf{R}$ can be seen as two interdependent random variables, and we hope to maximize their dependency on each other. To achieve this goal, we adopt Hilbert-Schmidt Independence Criterion (HSIC) \citep{gretton2005measuring}.

	HSIC acts as a non-parametric independence test between two random variables, $g_{\theta_X}(\mathbf{X})$ and $\mathbf{R}$, by computing the Hilbert-Schmidt norm of the covariance operator over the corresponding domains $\mathcal{G} \times \mathcal{R}$. Furthermore, let $k_g$ and $k_r$ be the kernels on $\mathcal{G}, \mathcal{R}$ with associated Reproducing Kernel Hilbert Spaces (RKHSs). A slightly biased empirical estimation of HSIC \citep{gretton2005measuring} could be written as follows:
	\begin{equation}
		\mathrm{HSIC}(\mathbf{S}, \mathbf{R}) = \frac{1}{(N-1)^{2}}\mathrm{tr}(\mathbf{HK_GH K_R }),
	\label{eq:HSIC}
	\end{equation}
	where $\mathbf{K_G} \in \mathbb{R}^{N\times N}$ with $\mathbf{K_G}_{ij} = k_g(x_i, x_j)$, $\mathbf{K_R} \in \mathbb{R}^{N\times N}$ with $\mathbf{K_R}_{ij} = k_r(y_i, y_j)$, and $\mathbf{H} \in \mathbb{R}^{N\times N}$ with $\mathbf{H}_{ij} = \mathbbm{1}_{\{i=j\}} -\frac{1}{(N-1)^{2}}$. In short, $\mathbf{K_G}$ and $\mathbf{K_R}$ respectively stand for the relationships between data and categories, and HSIC provides a statistical dependency guarantee on the learned embeddings and labels. 
}
\subsection{Kernel Learning via Deep Representation}
\label{subsec:DKL}
{
	Next, we explain how we construct the kernel $\mathbf{K_G}$ and $\mathbf{K_R}$.
	First of all, for simplicity, we adopt linear kernel for $k_g$:
	\begin{equation}
		k_g(x_i, x_j) = {g_{\theta_X}(x_i)}^{\top}\cdot g_{\theta_X}(x_j).
	\label{eq:kg}
	\end{equation}

	We incorporate multiple side information in $k_r$ as follows:
	\begin{equation}
		k_r(y_i, y_j) = \sum_{m=1}^{M}\frac{1}{M}\,k_{r^m}\left(y_i,y_j\right),
	\label{eq:kr}
	\end{equation}
	where $k_{r^m}\left(\cdot,\cdot\right)$ denotes the kernel choice for the $m_{th}$ side information $R^m$. We consider two variants of $k_{r^m}\left(\cdot,\cdot\right)$ based on whether $R^m$ is represented by class embeddings or tree-based label hierarchy.
	
	\subsubsection*{a) $R^m$ is represented by class embeddings:}

	Class embeddings can either be supervised features such as human annotated features or unsupervised features such as {\em word2vec} or {\em glove} features. Given $R^m = \{r_c^m\}_{c=1}^{C}$ with $r_c^m$ representing class embeddings of class $c$, we define $k_{r^m}\left(\cdot,\cdot\right)$ as:
	\begin{equation}
		k_{r^m}(y_i, y_j) = {f_{m,\theta_R}(r^m_{y_i})}^{\top}\cdot f_{m,\theta_R}(r^m_{y_j}),
	\label{eq:krt1}
	\end{equation}
	where $f_{m,\theta_R}(\cdot)$ denotes the non-linear mapping from $R^m$. 
	In this setting, we can capture the intrinsic structure by adjusting the categories' affinity through learning $f_{m,\theta_R}(\cdot)$ for different types of side information $R^m$.

	\subsubsection*{b) $R^m$ is represented by tree hierarchy:}

	If the labels form a tree hierarchy (e.g., {\em wordnet}~\citep{miller1995wordnet} tree structure in ImageNet), then we can represent the labels as a tree covariance matrix $\mathbf{B}$ defined in \cite{bravo2009estimating}, which is proved to be equivalent to the taxonomies in the tree \citep{blaschko2013taxonomic}. Specifically, following the definition of Theorem 2 in \cite{bravo2009estimating}, a matrix $\mathbf{B} \in \mathbb{R}^{C\times C}$ is the tree-structured covariance matrix if and only if $\mathbf{B} = \mathbf{V}\mathbf{D}\mathbf{V}^\top$ where $\mathbf{D} \in \mathbb{R}^{2C-1\times 2C-1}$ is the diagonal matrix indicating the branch lengths of the tree and $\mathbf{V} \in \mathbb{R}^{C \times 2C-1}$ denoting the topology. Please see Supplementary for the example of the covariance matrix for {\em Animals with Attributes} ($\mathsf{AwA}$) dataset \citep{lampert2014attribute}.

	For any given tree-based label hierarchy, we define $k_{r^m}\left(\cdot,\cdot\right)$ to be
	\begin{equation}
		k_{r^m}\left(y_i,y_j\right) = (\mathbf{B}^m)_{y_i,y_j} =  (\mathbf{Y^\top} \mathbf{B}^m\mathbf{Y})_{i,j}\,\,,
	\label{eq:krt2}
	\end{equation}
	where $\mathbf{Y} \in \{0,1\}^{C \times N} $ is the label matrix and $\mathbf{B}^m$ is the tree-structured covariance matrix of $R^m$.
	In other words, $k_{r^m}\left(y_i,y_j\right)$ indicates the weighted path from the root to the nearest common ancestor of nodes $y_i$ and $y_j$ (see Lemma 1 in \citep{blaschko2013taxonomic}).
	\vspace{2mm}

	Through the design in eq. \eqref{eq:kr}, we can try integrating different types of side information $R^m$ with both class-embedding and tree-hierarchy-structure representation. In short, maximizing eq. \eqref{eq:HSIC} makes the data representation kernel $\mathbf{K_G}$ maximally dependent on the side information $\mathbf{R}$ seen from the kernel matrix $\mathbf{K_R}$. Hence, introducing HSIC criterion provides an excellent way of transferring knowledge across different classes. Note that, if $\mathbf{K_R}$ is an identity matrix, then there are no relationships between categories, which results in a standard classification problem.

	So far, we have defined a joint learning on the support set~$\mathbf{S}$ and its side information $\mathbf{R}$. If we have access to different support set $\mathbf{S'}$ and the corresponding side information $\mathbf{R'}$, we can easily incorporate them into the HSIC criterion; i.e., $\mathrm{HSIC}(\{\mathbf{S}, \mathbf{S'}\}, \{\mathbf{R}, \mathbf{R'}\})$. Hence we can effectively transfer the knowledge both intra and inter sets.

}

	\begin{figure}[t!]
	\centering
	\includegraphics[width=0.33\textwidth]{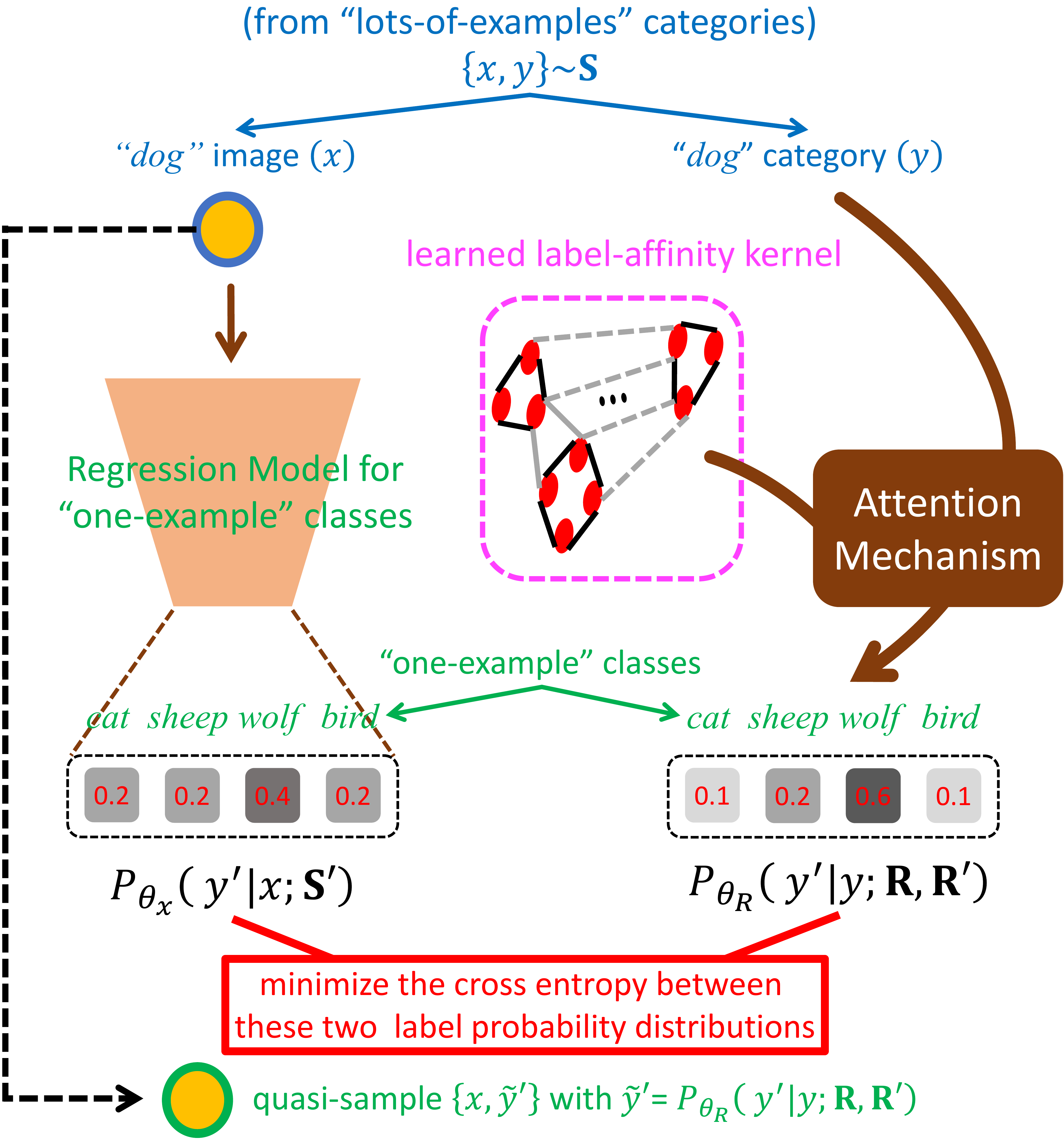}
	\vspace{-3mm}
	\caption{\footnotesize Quasi-samples generation: We take {\em dog} as an example class from ``lots-of-examples'' categories. ``One-example'' categories consist of {\em cat, sheep, wolf,} and {\em bird}. Best viewed in color.}
	\label{fig:quasi}
	\end{figure}

\subsection{Quasi-Samples Generation}
\label{subsec:set2set}
{
	Our second aim is to use a significant amount of data in `lots-of-examples' classes to learn the prediction model for `one-example' classes. We present an attention mechanism over 
the side information $\mathbf{R}$ and $\mathbf{R}'$ to achieve this goal.

	For a given data-label pair $\{x, y\}$ in $\mathbf{S}$, we define its quasi-label $\tilde{y}'$ as follows:
	\begin{equation}
		\tilde{y}' = P_{\theta_R}(y'|y; \mathbf{R}, \mathbf{R}') = \sum_{i\in \mathbf{S}'} a_r(y, y'_i) y'_i,
	\label{eq:quasilbl}
	\end{equation}
	where $a_r(\cdot,\cdot)$ acts as an attentional kernel from $\mathbf{R}$ to $\mathbf{R}'$, which can be formulated as 
	\begin{equation}
		a_r(y, y'_i) = \frac{e^{k_r(y, y'_i)}}{\sum_{j \in \mathbf{S}'}e^{k_r(y, y'_j)}}. 
	\label{eq:att_r}
	\end{equation}

	In other words, given the learned label affinity kernel, for each category in `lots-of-examples' classes, we can form a label probability distribution on the label space for `one-example' classes; i.e., $\tilde{y}' = P_{\theta_R}(y'|y; \mathbf{R}, \mathbf{R}')$. Moreover, given the other set $\mathbf{S'}$, we can also derive the label probability distribution $P_{\theta_X}(y'|x; \mathbf{S'})$ under any regression model (see Sec. \ref{subsec:pred}) for `one-example' classes. Our strategy is to minimize the cross entropy between $P_\theta(y'|x; \mathbf{S'})$ and $\tilde{y}'$. In short, we can treat each data-label pair \{$x,y$\} in `lots-of-examples' classes to be a quasi-sample \{$x, \tilde{y}'$\} for `one-example' classes, as illustrated in Fig.~\ref{fig:quasi}.
}

}

%% file: regre.tex
\section{REGRESSION AND TRAINING-TEST STRATEGY}
\label{sec:regre}
{

\subsection{Predictions by Regression}
\label{subsec:pred}
{
	We adopt Softmax (Parametric) Regression and Attentional (Non-Parametric) Regression to form the label probability distributions. Given the support set $\mathbf{S}$, we define the label prediction $\hat{y}$ to be
	\begin{equation}
		\hat{y} \coloneqq P_{\theta_X}({y}|{x}; \mathbf{S}).
	\label{eq:y_label_prob} 
	\end{equation}
	Due to the space limit, we elaborate two regression strategies in Supplementary.
}

\subsection{Training and Test Strategy - Learning in a One-Shot Setting}
\label{subsec:traing}
{

	Inspired by \cite{vinyals2016matching,ravi2017optimization}, we construct a training-time strategy to match the test-time evaluation strategy. 

	Let $T$ be the set of tasks defined on all possible label sets from `lots-of-examples' classes. Likewise, $T'$ is the set of tasks defined on all possible label sets from `one-example' classes. 
We first perform sampling from $T$ to $L$ and from $T'$ to $L'$ for choosing the tasks on the subsets of classes. Specifically, we force the number of classes in $L$ and $L'$ to be the number of `one-example' classes. For instance, if we randomly sample $5$ categories from `one-example' classes to perform an evaluation, we have $|L'| = 5$. Then, to match training and testing scenario, we also randomly sample $5$ categories from `lots-of-examples' classes so that $|L| = |L'|$ is achieved.

Next, we sample $\mathbf{S}$ along with the corresponding $\mathbf{R}$ from $L$ and sample $\mathbf{S'}$ along with the corresponding $\mathbf{R}'$ from $L'$. In order to strengthen the matching criterion between training and testing, we split $\mathbf{S}$ to $\mathbf{S}_{train}$ and $\mathbf{S}_{batch}$ ($\mathbf{S}_{train} \cup \mathbf{S}_{batch} = \mathbf{S}$ and $\mathbf{S}_{train} \cap \mathbf{S}_{batch} = \varnothing$). We have $|\mathbf{S}_{train}| = |\mathbf{S'}| = N'$ and also require $\mathbf{S}_{train}$ to have equal number of samples per category as in $\mathbf{S}'$.

	The first objective is to maximize the prediction of predicting labels in $\mathbf{S}_{batch}$, which can be formulated as 
	\begin{equation}
	\begin{split}
		O_1 = & E_{L \sim T}\bigg[E_{\mathbf{S}_{train}, \mathbf{S}_{batch} \sim L}\Big[ \\ 
				& \frac{1}{|\mathbf{S}_{batch}|}\sum_{i \in \mathbf{S}_{batch}} y_i^\top \,\mathrm{log} P_{\theta_X}\big(y_i|x_i; \mathbf{S}_{train}\big)\Big]\bigg].  
	\end{split}
	\label{eq:o1}
	\end{equation}
	Note that both $y_i$ and $P_{\theta_X}(y_i|x_i; \mathbf{S}_{train})$ are vectors of size~$\mathbb{
	R}^{C\times 1}$. 

	The second objective is to meet the HSIC criterion (eq. \eqref{eq:HSIC}) that maximally aligns the side information to the learned embeddings. We formulate the objective as follows:
	
	\begin{equation}
		\scalebox{0.85}{ $O_2 = E_{L \sim T; L' \sim T'}\bigg[E_{\mathbf{S}, \mathbf{R} \sim L; \mathbf{S'}, \mathbf{R}' \sim L'}\Big[  \mathrm{HSIC}\big(\{\mathbf{S}, \mathbf{S'}\}, \{\mathbf{R}, \mathbf{R'}\}\big)\Big]\bigg].$}
	\label{eq:o2}
	\end{equation}

	The third objective is to take the data in $\mathbf{S}_{batch}$ and their quasi-labels into consideration: namely, the data-label pairs $\{x_i, {\tilde{y}_i}'\}_{i=1}^{|\mathbf{S}_{batch}|}$, where ${\tilde{y}_i}'$ is defined in eq. \eqref{eq:quasilbl}. We maximize the negative cross entropy between ${\tilde{y}_i}'$ and the label probability distribution $P_{\theta_X}\big({y_i}'|x_i; \mathbf{S'}\big)$ in eq. \eqref{eq:y_label_prob}: 
	\begin{equation}
	\begin{split}
		O_3 = & E_{L \sim T; L' \sim T'}\bigg[E_{\mathbf{S}_{batch}, \mathbf{R} \sim L; \mathbf{S'}, \mathbf{R}' \sim L'}\Big[ \\ 
				& \frac{1}{|\mathbf{S}_{batch}|}\sum_{i \in \mathbf{S}_{batch}} {\tilde{y}_i}'^\top \,\mathrm{log} P_{\theta_X}\big({y_i}'|x_i; \mathbf{S'}\big)\Big]\bigg], 
	\end{split}
	\label{eq:o3}
	\end{equation}
	where both $\tilde{y}_i'$ and $P_{\theta_X}({y_i}'|x_i; \mathbf{S'})$ are of size $\mathbb{R}^{C'\times 1}$.

	The overall training objective is defined as follows:
	\begin{equation}
		\mathrm{max}\,\,\, O_1 + \alpha (O_2 + O_3), 
	\label{eq:cro_entro}
	\end{equation}
	where $\alpha$ is the trade-off parameter representing how we fuse side information to learn from `lots-of-examples' to `one-example' classes. We fix $\alpha = 0.1$ for simplicity in all of our experiments. We also perform fine-tuning over $S'$; that is, we update $\theta_X$ for a few iterations to maximize 
	\begin{equation}
		E_{L' \sim T'}
	\bigg[E_{\mathbf{S}'\sim L'}
	\Big[\frac{1}{|\mathbf{S}'|}\sum_{i \in \mathbf{S}'} {y'_i}^\top \,\mathrm{log} P_{\theta_X}\big(y'_i|x'_i; \mathbf{S}'\big)\Big]\bigg]. 
	\label{eq:finetun}
	\end{equation}

	Finally, for any given test example $x'_{test}$, the predicted output class is defined as
	\begin{equation}
		\hat{y}'_{test} = \mathrm{argmax}_{y'}\,P_{\theta_X}(y'|x'_{test}; \mathbf{S}'). 
	\label{eq:predtest}
	\end{equation}

}

}

%% file: experiments.tex
\section{EVALUATION}
\label{sec:exp}
{
	\begin{figure*}[t!]
	\vspace{-2mm}
	\centering
	\includegraphics[width=0.62\textwidth]{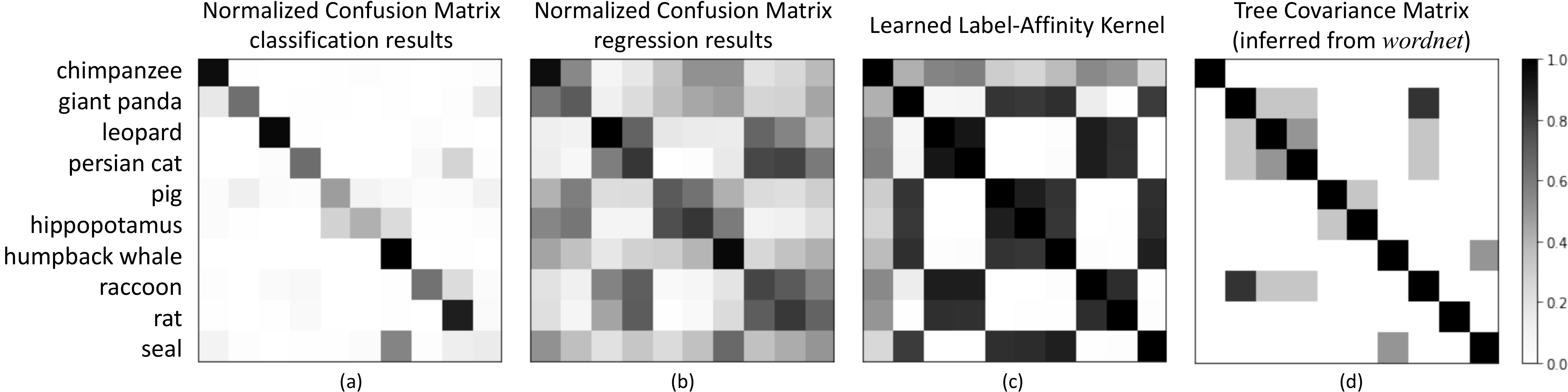}
	\vspace{-2mm}
	\caption{\footnotesize From left to right: (a) normalized confusion matrix for classification results, (b) normalized confusion matrix for regression results, (c) label-affinity kernel learned in $\mathsf{HSIC}$$_{attention}$, and (d) tree covariance matrix in Sec. \ref{subsec:DKL} inferred from {\em wordnet} for $\mathsf{AwA}$.}
	\label{fig:confusion}
	\vspace{-4mm}
	\end{figure*}

	In this Section, we evaluate our proposed method on top of two different networks (regression models): softmax regression ($\mathsf{softmax\_net}$) and attentional regression ($\mathsf{attention\_net}$). Attentional regression network can be 
viewed as a variant of Matching Networks \citep{vinyals2016matching} without considering the Fully Conditional Embeddings (FCE) in \citep{vinyals2016matching}. In our experiments, two datasets are adopted for one-shot recognition task: Caltech-UCSD Birds 200-2011 ($\mathsf{CUB}$) \citep{WelinderEtal2010} and Animals with Attributes ($\mathsf{AwA}$) \citep{lampert2014attribute}. $\mathsf{CUB}$ is a fine-grained dataset containing bird species where its categories are both visually and semantically similar, while $\mathsf{AwA}$ is a general dataset which contains animal species across land, sea, and air. We use the same training+validation/ test splits in \cite{akata2015evaluation,tsai2017learning}: $150$/$50$ classes for $\mathsf{CUB}$ and $40$/$10$ classes for $\mathsf{AwA}$.

	We consider four types of side information: supervised human annotated attributes ($\mathit{att}$) \citep{lampert2014attribute}, unsupervised Word2Vec features ($\mathit{w2v}$) \citep{mikolov2013distributed}, unsupervised Glove features ($\mathit{glo}$) \citep{pennington2014glove}, and the tree hierarchy ($\mathit{hie}$) inferred from {\em wordnet} \citep{miller1995wordnet}. 
Human annotated attributes $\mathit{att}$ are represented as $312$-/$85$-dimensional features for $\mathsf{CUB}$ and $\mathsf{AwA}$, respectively. $\mathit{w2v}$ and $\mathit{glo}$ are $400$-dimensional features pre-extracted from Wikipedia provided by \citep{akata2015evaluation}. On the other hand, $\mathit{hie}$ are not represented as feature vectors but define the hierarchical relationships between categories. Please see Appendix for the tree hierarchy of $\mathsf{CUB}$ and $\mathsf{AwA}$. The implementation details are also provided in Appendix. 
We report results averaged over $40$ random trials.
\subsection{One-Shot Recognition}
\label{subsec:one_shot}
{

	\begin{table}[t]
	\vspace{-2mm}
	\footnotesize
	\centering
	\caption{\footnotesize Average performance for standard one-shot recognition task. Our proposed methods jointly learn with all four side information: $\mathit{att}$, $\mathit{w2v}$, $\mathit{glo}$, and $\mathit{hie}$.}
	\vspace{1mm}
	\label{tbl:one_shot}
	\scalebox{0.7}
	{
	\begin{tabular}{|c||c|c|}
	\hline
	network / Dataset             & $\mathsf{CUB}$           & $\mathsf{AwA}$    \\ \hline \hline
	$\mathsf{softmax\_net}$    &   26.93    $\pm$   2.41     &    66.39    $\pm$   5.38   \\ 
	$\mathsf{HSIC}$$_{softmax}^\dagger$            &    29.26    $\pm$    2.22   &   69.98   $\pm$    5.47    \\ 
	$\mathsf{HSIC}$$_{softmax}$            &    31.49   $\pm$   2.28     &    71.29    $\pm$  5.64    \\ \hline \hline
	$\mathsf{attention\_net}$ [\cite{vinyals2016matching}]                  &    29.12   $\pm$    2.44    &    72.27    $\pm$   5.82   \\ 
	$\mathsf{HSIC}$$_{attention}^\dagger$            &    33.12    $\pm$   2.48   &   {\bf 77.86     $\pm$    4.76}  \\
	$\mathsf{HSIC}$$_{attention}$            &    {\bf 33.75  $\pm$   2.43}      &   76.98    $\pm$   4.99    \\ \hline
	\end{tabular}
	}
	\vspace{-3mm}
	\end{table}

	\begin{figure*}[t!]
	\centering
	\includegraphics[width=0.75\textwidth]{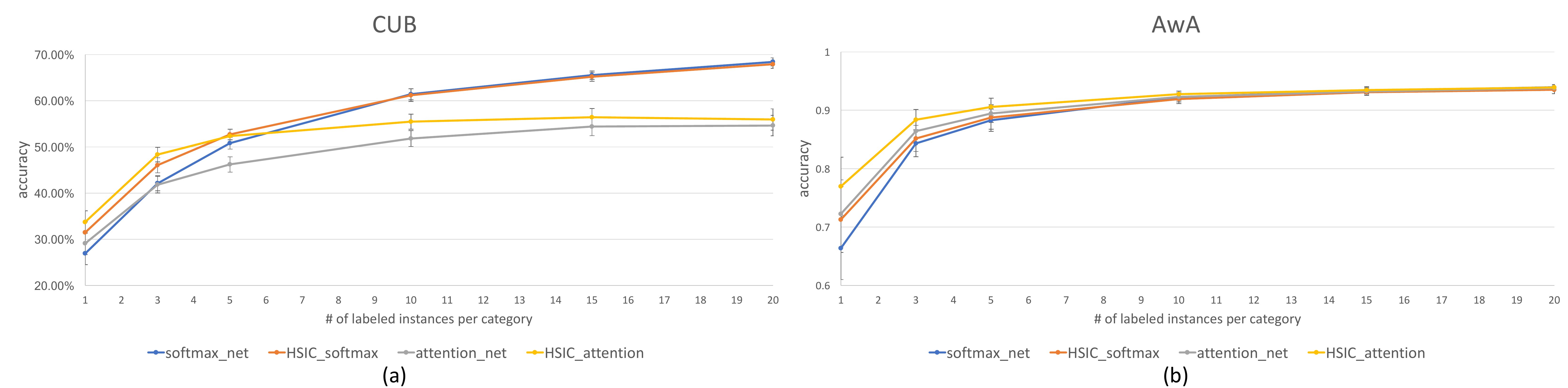}
	\vspace{-4mm}
	\caption{\footnotesize Experiment for increasing labeled instance per category in test classes. Our proposed methods jointly learn with all four side information: $\mathit{att}$, $\mathit{w2v}$, $\mathit{glo}$, and $\mathit{hie}$. Best viewed in color.}
	\label{fig:incre_lbl}
	\vspace{-3mm}
	\end{figure*}

	First, we perform one-shot recognition tasks on $\mathsf{CUB}$ and $\mathsf{AwA}$: for test classes, only one labeled instance is provided during training and the rest of the instances are for prediction in test time. We denote our proposed method using softmax regression and attentional regression as $\mathsf{HSIC}$$_{softmax}$ and $\mathsf{HSIC}$$_{attention}$, respectively. $\mathsf{HSIC}$$_{softmax}$ and $\mathsf{HSIC}$$_{attention}$ relax to $\mathsf{softmax\_net}$ and $\mathsf{attention\_net}$ when we only consider the objective $O_1$ ($\alpha = 0$) in eq. \eqref{eq:cro_entro}. To complete our experiments, we provide two more variants: $\mathsf{HSIC}$$_{softmax}^\dagger$ and $\mathsf{HSIC}$$_{attention}^\dagger$. They stand for our proposed method without considering $O_3$ in eq. \eqref{eq:cro_entro}; that is, we do not generate quasi-samples for our test classes (`one-example' ones) from instances in training classes (`lots-of-examples' ones). The results are reported using top-1 classification accuracy (\%) from eq.~\eqref{eq:predtest} on test samples in test classes.

	{\bf Experiments:} Table \ref{tbl:one_shot} lists the average recognition performance for our standard one-shot recognition experiments. $\mathsf{HSIC}$$_{softmax}$ and $\mathsf{HSIC}$$_{attention}$ are jointly learned with all four types of side information: $\mathit{att}$, $\mathit{w2v}$, $\mathit{glo}$, and $\mathit{hie}$. We first observe that all methods perform better on $\mathsf{AwA}$ than in $\mathsf{CUB}$ dataset. This is primarily because $\mathsf{CUB}$ is a fine-grained dataset where inter-class differences are very small, which increases its difficulty for object classification. 
Moreover, the methods with side information achieve superior performance over the methods which do not learn with side information. For example, $\mathsf{HSIC}$$_{softmax}$ improves over $\mathsf{softmax\_net}$ by $4.56\%$ on $\mathsf{CUB}$ dataset and $\mathsf{HSIC}$$_{attention}$ enjoys $4.71\%$ gain over $\mathsf{attention\_net}$ on $\mathsf{AwA}$ dataset. These results indicate that fusing side information can benefit one-shot learning. 

	Next, we examine the variants of our proposed architecture. In most cases, the construction of the quasi-samples benefits the one-shot learning. The only exception is the $0.88\%$ performance drop from $\mathsf{HSIC}$$_{attention}^\dagger$ to $\mathsf{HSIC}$$_{attention}$ in $\mathsf{AwA}$. Nevertheless, we find that our model converges faster when introducing the technique of generating quasi-samples. 

Finally, methods based on {\em attentional regression} have better performance over methods using {\em softmax regression}. For instance, we find $2.19\%$ performance deterioration from $\mathsf{attention\_net}$ to $\mathsf{softmax\_net}$ in $\mathsf{CUB}$ and $5.69\%$ performance improvement from $\mathsf{HSIC}$$_{softmax}$ to $\mathsf{HSIC}$$_{attention}$ in $\mathsf{AwA}$. The non-parametric characteristic of {\em attentional regression} enables the model to learn fewer parameters (compared to {\em softmax regression}) and enjoys better performance in one-shot setting.

	\begin{figure}[t!]
	\vspace{-1mm}
	\centering
	\includegraphics[width=0.38\textwidth]{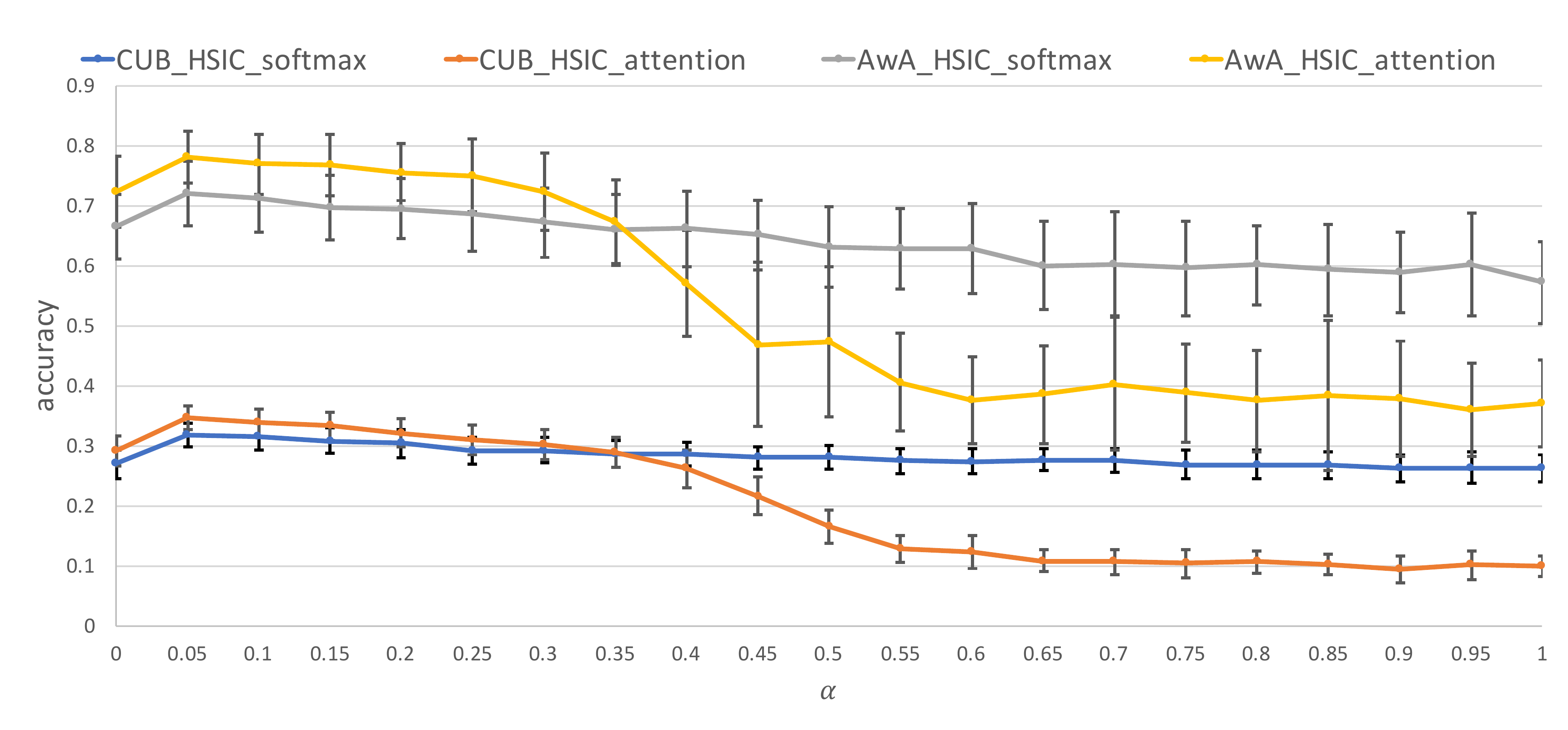}
	\vspace{-4mm}
	\caption{\footnotesize Parameter sensitivity analysis experiment. Our proposed methods jointly learn with all four side information: $\mathit{att}$, $\mathit{w2v}$, $\mathit{glo}$, and $\mathit{hie}$. Best viewed in color.}
	\label{fig:alpha}
	\vspace{-4mm}
	\end{figure}

	{\bf Confusion Matrix and the Learned Class-Affinity Kernel:} Following the above experimental setting, for test classes in $\mathsf{AwA}$, in Fig. \ref{fig:confusion}, we provide the confusion matrix, the learned label-affinity kernel using $\mathsf{HSIC}$$_{attention}$, and the tree covariance matrix \citep{bravo2009estimating}. We first take a look at the normalized confusion matrix for classification results. For example, we observe that {\em seal} is often misclassified as {\em humpback whale}; and from the tree covariance matrix, we know that {\em seal} is semantically most similar to {\em humpback whale}. Therefore, even though our model cannot predict {\em seal} images correctly, it still can find its semantically most similar classes. 

	Additionally, it is not surprising that Fig.~\ref{fig:confusion}(b), normalized confusion matrix, is visually similar to Fig.~\ref{fig:confusion}(c), the learned class-affinity kernel. 
The reason is that one of our objectives is to learn the output embeddings of images to be maximally dependent on the given side information. Note that, in this experiment, our side information contains supervised human annotated attributes, unsupervised word vectors (Word2Vec \citep{mikolov2013distributed} and Glove \citep{pennington2014glove}), and a {\em WordNet} \citep{miller1995wordnet} tree hierarchy.

	On the other hand, we also observe the obvious change in classes relationships from {\em WordNet} tree hierarchy (Fig. \ref{fig:confusion} (d)) to our learned class-affinity kernel (Fig. \ref{fig:confusion} (c)). For instance, {\em raccoon} and {\em giant panda} are species-related, but they distinctly differ in size and color. This important information is missed in {\em WordNet} but not missed in human annotated features or word vectors extracted from Wikipedia. Hence, our model bears the capability of arranging and properly fusing various types of side information.

	{\bf Parameter Sensitivity on $\alpha$:} Since $\alpha$ stands for the trade-off parameter for fusing side information through HSIC and quasi-examples generation technique, we studied how it affects model performance. We alter $\alpha$ from $0$ to $1.0$ by step size of $0.05$ for both $\mathsf{HSIC}$$_{softmax}$ and $\mathsf{HSIC}$$_{attention}$ models. 
Fig.~\ref{fig:alpha} shows that larger values of $\alpha$ does not lead to better performance. When $\alpha \leq 0.3$, our proposed method outperforms $\mathsf{softmax\_net}$ and $\mathsf{attention\_net}$. Note that $\mathsf{HSIC}$$_{softmax}$ and $\mathsf{HSIC}$$_{attention}$ relax to $\mathsf{softmax\_net}$ and $\mathsf{attention\_net}$ when $\alpha = 0$. When $\alpha > 0.3$, the performance of our proposed method begins to drop significantly, especially for $\mathsf{HSIC}$$_{attention}$. This is primarily because too large values of $\alpha$ may cause the output embeddings of images to be confused by semantically similar but visually different classes in the learned label-affinity kernel (e.g., Fig. \ref{fig:confusion} (c)).

	{\bf From One-Shot to Few-Shot Learning:} Next, in Fig. \ref{fig:incre_lbl}, we increase the labeled instances in test classes and evaluate the performance of $\mathsf{softmax\_net}$, $\mathsf{attention\_net}$, and our proposed architecture. We randomly label $1$ (one-shot setting), $3$, $5$, $10$, $15$, and $20$ (few-shot setting) instances in test classes. These labeled instances are used for training, while the rest unlabeled instances are used for prediction 
at the test stage. We observe that $\mathsf{HSIC}$$_{softmax}$ converges to $\mathsf{softmax\_net}$ and $\mathsf{HSIC}$$_{attention}$ converges to $\mathsf{attention\_net}$ when more labeled data are available in test classes during training. In other words, as labeled instances increase, the power of fusing side information within deep learning diminishes. This result is quite intuitive as deep architecture perform 
well when training on lots of labeled data.

	For the fine-grained dataset $\mathsf{CUB}$, we also observe that {\em attentional regression} methods are at first outperform {\em softmax regression} methods, but perform worse when more labeled data are present during training. Recall that, in setting, {\em softmax regression} networks have one additional softmax layer (one-hidden-layer fully-connected neural network) compared to {\em attentional regression} networks. Therefore, {\em softmax regression} networks can deal with more complex regression functions (i.e., regression for the fine-grained $\mathsf{CUB}$ dataset) as long as they have enough labeled examples. 

	{\bf More Experiments and Comparisons:} Due to space limit, we leave more experiments and comparisons in Supplementary. First, we provide the experiments on the availability of various types of side information. Second, we provide the experiments for comparing the proposed method with direct side information fusion and ReViSE \citep{tsai2017learning}. Last, we also provide the experiments for expanding training- and test-time categories search space.

}